\documentclass[utf8]{frontiersSCNS} 

\usepackage{url,hyperref,lineno,microtype,subcaption}
\usepackage[onehalfspacing]{setspace}
\usepackage{multirow}
\usepackage{amsmath} 
\usepackage{mathtools}
\usepackage{color}
\usepackage{algorithm}
\usepackage{algorithmic}

\def\keyFont{\fontsize{8}{11}\helveticabold }
\def\firstAuthorLast{Huan Yin {et~al.}} 
\def\Authors{Huan Yin\,$^{1}$, Xuecheng Xu\,$^{1}$, Yue Wang\,$^{1,*}$ and Rong Xiong\,$^{1}$}
\def\Address{$^{1}$Institute of Cyber-Systems and Control, College of Control Science and Engineering, Zhejiang University, Hangzhou, China
}
\def\corrAuthor{Yue Wang}

\def\corrEmail{ywang24@zju.edu.cn}

\begin{document}
\onecolumn
\firstpage{1}

\title[Radar-to-Lidar: Heterogeneous Place Recognition]{Radar-to-Lidar: Heterogeneous Place Recognition via Joint Learning} 

\author[\firstAuthorLast ]{\Authors} 
\address{} 
\correspondance{} 

\extraAuth{}

\maketitle

\begin{abstract}


Place recognition is critical for both offline mapping and online localization. However, current single-sensor based place recognition still remains challenging in adverse conditions. In this paper, a heterogeneous measurements based framework is proposed for long-term place recognition, which retrieves the query radar scans from the existing lidar maps. To achieve this, a deep neural network is built with joint training in the learning stage, and then in the testing stage, shared embeddings of radar and lidar are extracted for heterogeneous place recognition. To validate the effectiveness of the proposed method, we conduct tests and generalization experiments on the multi-session public datasets compared to other competitive methods. The experimental results indicate that our model is able to perform multiple place recognitions: lidar-to-lidar, radar-to-radar and radar-to-lidar, while the learned model is trained only once. We also release the source
code publicly: \href{https://github.com/ZJUYH/radar-to-lidar-place-recognition}{https://github.com/ZJUYH/radar-to-lidar-place-recognition}.

\tiny
 \keyFont{ \section{Keywords:} radar, lidar, heterogeneous measurements, place recognition, deep neural network} 
\end{abstract}

\section{Introduction}

Place recognition is a basic technique for both field robots in the wild and automated vehicles on the road, which helps the agent to recognize revisited places when travelling. In the mapping session or Simultaneous Localization and Mapping (SLAM), place recognition is equal to loop closure detection, which is indispensable for global consistent map construction. In the localization session, place recognition is able to localize the robot via data retrieval, thus achieving global localization from scratch in GPS-denied environments.

\begin{figure}[t]
	\begin{center}
		\includegraphics[width=\linewidth]{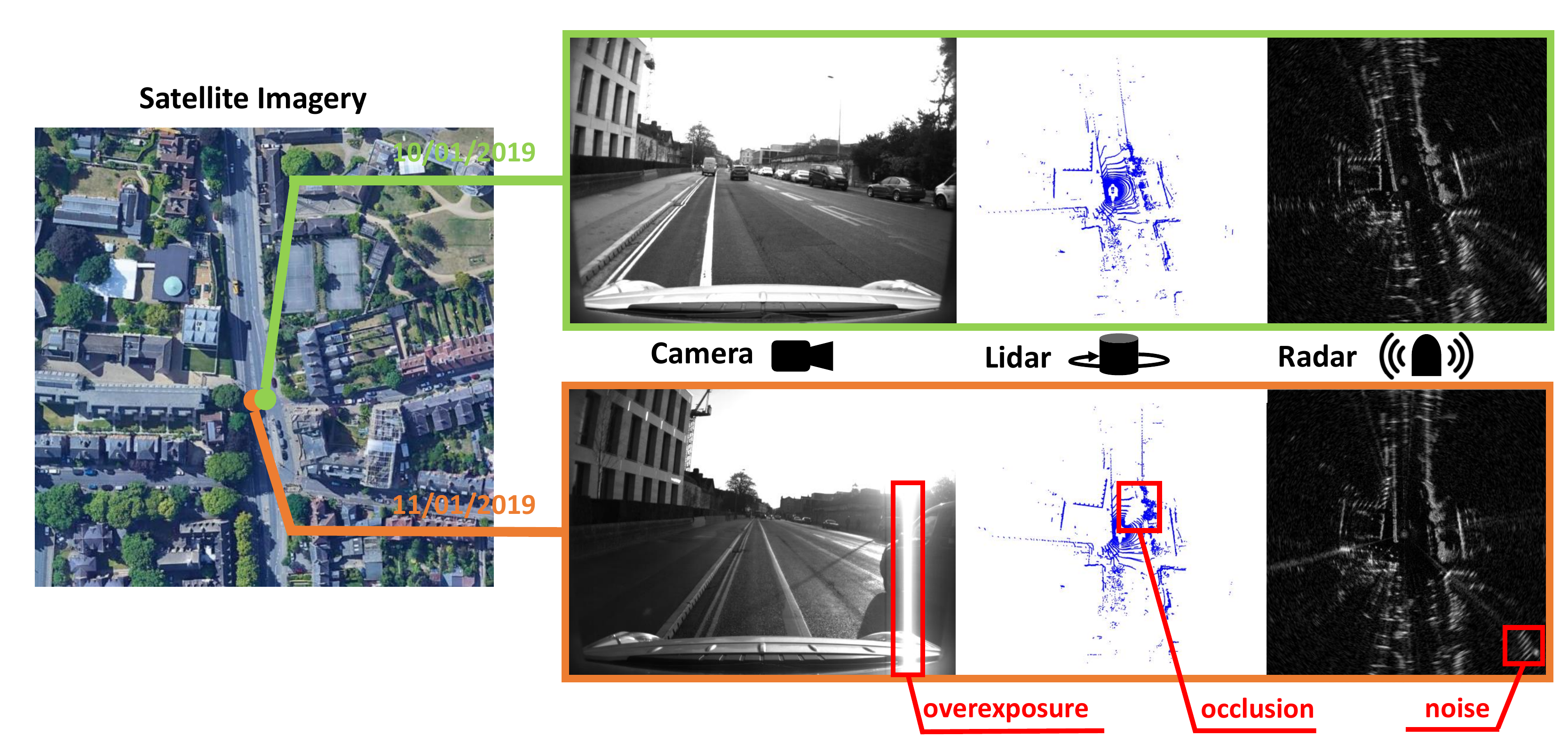}
	\end{center}
	\caption{ The sensor data collected at the same place but different time. These data are seleceted from the Oxford Radar RobotCar dataset. Obivously, every sensor has its weakness for long-term robotic perception. }
	\label{fig:sensors}
\end{figure}

Essentially, the major challenge for place recognition is how to return the correct place retrieval under the environmental variations. For visual place recognition \cite{lowry2015visual}, the illumination change is the considerable variation across day and night, which makes the image retrieval extremely challenging for the mobile robots. As for lidar-based perception \cite{elhousni2020survey}, the lidar scanner does not suffer from the illumination variations, and provides precise measurements of the surrounding environments. But in adverse conditions, fog and strong light etc., or in highly dynamic environments, the lidar data are affected badly by low reflections \cite{carballo2020libre} or occlusions \cite{kim2020mulran}. Compared to the vision or lidar, radar sensor is naturally lighting and weather invariant, and has been widely applied in the Advanced Driver Assistance Systems (ADAS) for object detection. But on the other hand, radar sensor generates noisy measurements, thus resulting in challenges for radar-based place recognition. So overall, there still remain different problems in the conventional single-sensor based place recognition, and we present a case study in Figure \ref{fig:sensors} for understanding.


Intuitively, these problems arise from the sensor itself at the front-end, not the recognition algorithm at the back-end. To overcome these difficulties, we consider a combination of multiple sensor is desired for long-term place recognition, for example, building map database in stable environments, while performing query-based place recognition in adverse conditions. One step further, given that large-scale high-definition lidar maps have been deployed for commercial use \cite{li2017overview}, a radar-to-lidar based place recognition is a feasible solution, which is robust to the weather changes and does not require extra radar mapping session, thus making the place recognition module more applicable in the real world. 

In this paper, we propose a heterogeneous place recognition framework using joint learning strategy. Specifically, we first build a shared network to extract feature embedings of radar and lidar, and then rotation-invariant signatures are generated via fourier transform. The whole network is trained jointly with the heterogeneous measurement inputs. In the experimental section, the trained model achieves not only homogeneous place recognition for radar or lidar, but also the heterogeneous task for radar-to-lidar. In summary, the contributions of this paper is listed as follows:
\begin{itemize}
	\item A deep neural network is proposed to extract the feature embeddings of radar and lidar, which is trained with a joint triplet loss. The learned model is trained once and achieves multiple place recognition tasks.
	\item We conduct the multi-session experiments in two public datasets, also with the comparisons to other methods, thus demonstrating the effectiveness of the proposed method in the real-world application.
\end{itemize}
The rest of this paper is organized as follows: Section \ref{related_works} reviews the related works. Our proposed method is
introduced in Section \ref{methods}. We conduct the experiments using two public datasets in Section \ref{experiments}. Finally, we conclude a brief overview of our method and a future outlook in Section \ref{conclusion}.

\section{Related Works}
\label{related_works}

\textbf{Visual-Based Place Recognition} Visual place recognition aims at retrieving the similar images in the database according to the current image and robot pose. Various image features have been proposed to measure the image similarities, SURF \cite{bay2006surf} and ORB \cite{rublee2011orb} etc., and also on the pre-defined patches \cite{filliat2007visual, jegou2010aggregating}. Based on these front-end descriptors, some researchers proposed probabilistic approach to \cite{cummins2008fab}, or searched the best candidates in the past sequences \cite{milford2012seqslam}. But due to the limitation of handcrafted descriptors, these visual place recognition methods are sensitive to the environmental changings. 

With the increasing development of deep learning technique, more researchers built Convolutional Neural Networks (CNN) to solve the visual place recognition problem. Compared to the conventational descriptors, the CNN based methods are more flexible on trainable parameters \cite{arandjelovic2016netvlad}, and also more robust across the season changes \cite{latif2018addressing}. Currently, there are some open issues to be studied in the vision and robotics community, such as feature selection and fusion for visual place recognition \cite{zhang2020visual}.

\textbf{Lidar-Based  Place Recognition} According to the generation process of representations, the lidar-based place recognition methods can also be classified as two categories, the handcrafted based and the learning based respectively. Bosse and Zlot \cite{bosse2013place} proposed to extract the 3D keypoints from 3D lidar points and performed the place recognition via keypoint voting. In \cite{dube2020segmap}, local segments were learned from CNN to represent places effectively. Despite the local representations, several global handcrafted descriptors \cite{he2016m2dp, kim2018scan, wang2019lidar} or matching based methods \cite{gentil2020gaussian} were proposed to solve point-cloud-based place recognition. These global descriptors are based on the structure of the range distribution and efficient for place recognition. Similarly, some learning based methods first generated the representations according to the statistical properties, then fed them into into the classifiers \cite{granstrom2011learning} or CNN \cite{yin20193d, chen2020overlapnet}. In addition, some researchers proposed to learn the point features in an end-to-end manner recently \cite{uy2018pointnetvlad, liu2019lpd}, while these methods bring more complexity for network training and recognition inference. 

\textbf{Radar-Based Mapping and Localization} Compared to the cameras and laser scanners, radar sensor has already been used in automotive industry \cite{krstanovic2012radar}. With the development of Frequency-Modulated Continuous-Wave (FMCW) radar sensor \footnote{\href{https://navtechradar.com}{https://navtechradar.com}}, the mapping and localization topics are studied in the recent years, for example the RadarSLAM \cite{hong2020radarslam}, radar odometry \cite{cen2018precise, barnes2020masking} and radar localization on lidar maps \cite{yin2020radar,yin2021rall}. 

For radar-based place recognition, Kim et al. \cite{kim2020mulran} extended the lidar-based handcrafted representation \cite{kim2018scan} to radar data directly. In \cite{suaftescu2020kidnapped}, NetVLAD \cite{arandjelovic2016netvlad} was used to achieve radar-to-radar place recognition. Then, the researchers used sequential radar scans to improve the localization performance \cite{gadd2020look}. In this paper, a deep neural network is also proposed to extract feature embeddings, but the proposed framework aims at heterogeneous place recognition.

\textbf{Multi-Modal Measurements for Robotic Perception} Many mobile robots and vehicles are equipped with multiple sensors and various perception tasks can be achieved via heterogeneous sensor measurements, for example, visual localization on point cloud maps \cite{ding2019persistent,feng20192d3d} and radar data matching on satellite images \cite{tang2020rsl}. While for place recognition, there are few methods performed on heterogeneous measurements. Cattaneo et al. \cite{cattaneo2020global} built shared embedding space for visual and lidar, thus achieving global visual localization on lidar maps via place recognition. Some researchers proposed to conduct the fusion of image and lidar points for place recognition \cite{xie2020large}. Similarly, in \cite{pan2020coral}, the authors first built local dense lidar maps from raw lidar scans, and then proposed a compound network to align the feature embeddings of image and lidar maps. The proposed framework was able to achieve bi-modal place recognition using one global descriptor. In summary, we consider the matching or fusion of multi-modal measurements is a growing trend in robotics community.

\section{Methods}
\label{methods}

\begin{figure}[t]
	\begin{center}
		\includegraphics[width=\linewidth]{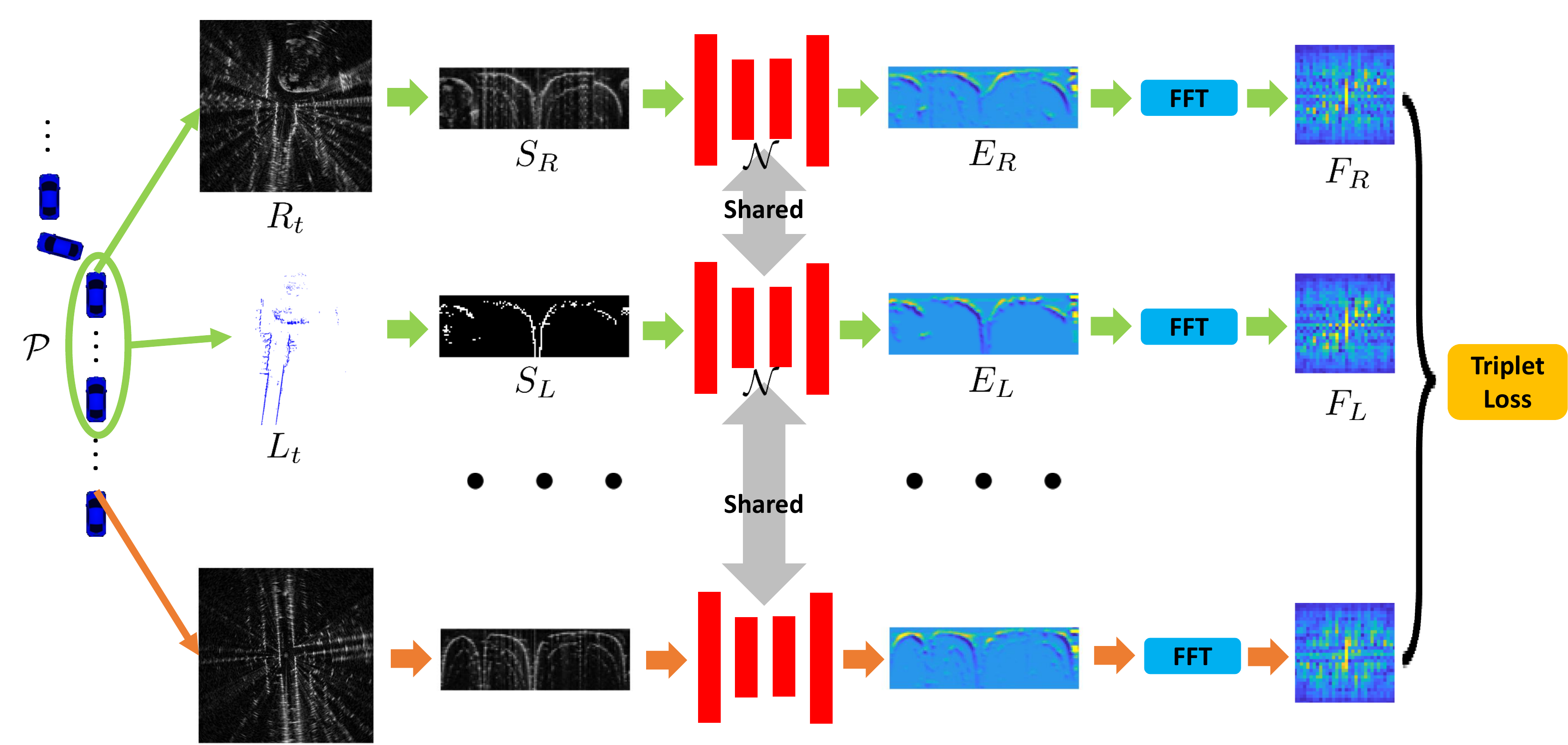}
	\end{center}
	\caption{ Our proposed framework to train the joint place recognition for radar and lidar data. The first and second row indicate that the radar scan and lidar submap are collected from the same place, while for the last row, it is regarded as a negative sample in the learning stage.}
	\label{fig:network}
\end{figure}

Our proposed framework is presented in Figure \ref{fig:network}. There are several pipelines, including building lidar submaps, generation of the learned signatures etc. Finally, the learned model generates low-dimensional representations for place recognition task in this paper.  

\subsection{Building Lidar Submaps}

Generally, the detection range of radar is much longer than lidar. To reduce the data difference for joint learning, we set the maximum range as $r_{max}$ meters in radar and lidar data. The 3D lidar also contains more height information compared to the 2D radar, and therefore we remove the redundant laser points and only keep the points near the radar sensor on Z-axis.

Despite the above difference, the lidar points are more sparse and easily occuluded by other onboard sensors, and also by the dynamics on road. In this paper, we follow the experimental settings in \cite{pan2020coral}, and propose to build submaps by accumulating the sequential lidar data. Then the problem turns to how many poses or lidar scans should be used for submap building. Suppose the robot travels at a pose $p_t$, and we use the poses from the start pose $p_{t-t_1}$ to the end pose $p_{t+t_2}$, where $t$, $t_1$ and $t_2$ are time indexes. In order to keep the consistency of lidar submap and radar scan, we propose to achieve the $t_1$ and $t_2$ using the following criteria:
\begin{itemize}
\item The maximum euclidean distance from $p_{t-t_1}$ to $p_t$ is limited to $r_{max}$ meters, guaranteeing the travelled length of the submap. It is same for the maximum distance from $p_t$ to $p_{t+t_2}$.
\item The maximum rotational angle from $p_{t-t_1}$ to $p_t$ should not be greater than $\theta_{max}$, which makes the lidar submaps more feasible at the turning corners. The rotational angles for mobile robots and vehicles are usually the yaw angles on the ground plane. Also, it is same for the maximum yaw angle from $p_t$ to $p_{t+t_2}$.
\end{itemize}

On the other hand, the lidar submap is desired to be as long as radar scan. Based on the criteria above, we can fomulate the retrieval of $t_1$ as follows:
\begin{equation} \label{submap_start}
\begin{aligned}
& \underset{t_1}{\text{minimize}}
& & t-t_1 \\
& \text{subject to}
& & \left\|p_t - p_{t-t_1}\right\|_2 \leq r_{max} \\
&&& \left| \angle p_t - \angle p_{t-t_1}\right|  \leq \theta_{max} .
\end{aligned}
\end{equation}
where $\angle$ is the yaw angle of pose. It is maximize operation for $t+t_2$ to achieve the $p_{t+t_2}$ with similar constraints. Specifically, we use a greedy strategy to search $t_1$ and $t_2$, and we use the retrieval of $t_1$ as an example in Algorithm~\ref{algorithm}. With obtained boundaries $t-t_1$ and $t+t_2$, a lidar submap can be built directly by accumulating the lidar scans from $p_{t-t_1}$ to $p_{t+t_2}$, thus achieving a lidar submap at the robot pose $p_t$. Note that we use ground truth poses for the map building in this subsection.

One might suggest that radar mapping should also be considered. However, as shown in Figure~\ref{fig:sensors}, there exist false positives and other noises in radar scans, resulting in inapplicable representations after radar mapping. In this context, we propose to build lidar submaps rather than radar submaps for heterogeneous place recognition in this paper.

\begin{algorithm}[t]
	\caption{Greedy search for $t_1$ retrieval}
	\label{algorithm}
	\begin{algorithmic}
		\REQUIRE ~~\\ 
		The robot pose: $t$; \\
		The maximum euclidean distance: $r_{max}$; \\
		The maximum rotational angle: $\theta_{max}$; \\
		\ENSURE
		\STATE $t_1 = 1$ 
		\WHILE{$\left\|p_t - p_{t-t_1}\right\|_2 \leq r_{max}$ \\ and $\left| \angle p_t - \angle p_{t-t_1}\right|  \leq \theta_{max}$}
		\STATE $t_1=t_1+1$
		\ENDWHILE
	\end{algorithmic}
\end{algorithm}

\subsection{Signature Generation}

With the built lidar submaps $L_t$ and the radar scan $R_t$ collected at the same pose, we first use the ScanContext \cite{kim2018scan, kim2020mulran} to extract the representations $S_L$ and $S_R$ respectively. Specifically, for radar scan, the ScanContext is the polar representation essentially. As for the lidar point clouds, we follow the settings in our previous research \cite{xu2020disco}, in which occupied representation achieves the best performance. Since the radar data are generated on 2D $x-y$ plane, we use the single-layer binary grids as the occupied representation, which indicates that the height information is removed in this paper. 

Then we build network $\mathcal{N}$ to extract the feature embeddings $E_L$ and $E_R$. Specifically, $S_L$ and $S_R$ are fed into a shared U-Net architecture \cite{ronneberger2015u}, and our hidden representations $E_L$ and $E_R$ are obtained in the feature space via feed-forward network. One might suggest that the networks should be different for the heterogeneous measurements, but considering that there are commons between radar and lidar, we propose to use the siamese structure to extract the embeddings. We validate this structure in following experimental section.

Finally, we follow the process in our previous work \cite{xu2020disco} and apply Fast Fourier Transformation (FFT) to the polar BEV (bird's eye view) representation. To make the process more efficient, we only extract the informative low-frequency component using low-pass filter, and then signatures $F_L$ and $F_R$ are generated in the frequency domain. Theoretically, the rotation of vehicle equals to the translation of $E_L$ and $E_R$ in polar domain, and the magnitude of frequency spectrum is translation-invariant actually, thus making the final signatures rotation-invariant to the vehicle heading. Overall, we summarize the processes of the signature generation as follows:
\begin{equation} \label{process}
L_t,R_t  \xrightarrow[]{\text{ScanContext}} S_L,S_R  \xrightarrow[]{\mathcal{N}} E_L, E_R \xrightarrow[]{\text{FFT}} F_L, F_R
\end{equation}
and the visualization of Equation \ref{process} is presented in Figure \ref{fig:network}. Essentially, the final signatures $F$ are the learned 'fingerprints' of places. If two signatures are similar, the two places are close to each other and vice versa.

\subsection{Joint Training}

With the generated batch $\mathcal{F}=\{F_L, F_R\}$, we propose to achieve the heterogeneous place recognition using joint training strategy. Specifically, radar-to-radar (R2R), lidar-to-lidar (L2L) and radar-to-lidar (R2L) are trained together under the supervision of one loss function. To achieve this, we build the triplet loss, and mix all the combinations in it, which is formulated as follows:
\begin{equation} \label{loss}
\mathcal{L}_1 = \frac{1}{\left| \mathcal{F} \right|} \sum_{F\in\mathcal{F}} \text{max}(0, m + \text{pos}(F) - \text{neg}(F))
\end{equation}
where $m$ is a margin value. $F$ is any combination in the set $\mathcal{F}$, for example, the combination of $\{ \text{anchor}, \text{positive}, \text{negative} \}$ samples can be $\{ \text{radar}, \text{lidar}, \text{radar} \}$, or $\{ \text{radar}, \text{radar}, \text{lidar} \}$. The number of these combinations for training is $\left| \mathcal{F} \right| = 2^3$. $\text{pos}(F)$ is the measured Euclidean distance for a pair of positive samples, while $\text{neg}(F)$ for the anchor and negative sample.

In this way, the trained model achieves not only the single-sensor based place recognition, but also the homogeneous task for radar-to-lidar. Note that there are three place recognition tasks in this paper, R2R, L2L and R2L (or L2R), but we only train the network once.

\section{Experiments}
\label{experiments}

The experiments are introduced in this section. We first present the setup and configuration, then followed the loop closure detection and place recognition results, with the comparison to other methods. Case study examples are also included for better understanding of the result. Considering that most of the existing maps are built by lidar in robotics community, radar-to-lidar task is more interesting and meaningful for heterogeneous place recognition, compared to lidar-to-radar task. Therefore, we only perform radar-to-lidar to demonstrate the effectiveness of our proposed framework.

\subsection{Implementation and Experimental Setup}
\label{experiments_setup}

The proposed network is implemented using PyTorch \cite{paszke2019pytorch}. We set the maximum range distance as $r_{max}=80m$, and set $\theta_{max}=90^\circ$. For the ScanContext representation, we set the size as $40 \times 120$, and achieve the $32\times32$ low-frequency signature finally. Some parameters have an influence on the model performance, for example, bin sizes, and we follow the experimental settings in \cite{xu2020disco}, which have been demonstrated to be effective and efficient. In the training session, there are more than 8000 samples generated randomly and the batchsize is set as 16. We run 6 epochs with the Adam optimizer \cite{kingma2014adam} and a decayed learning rate from 0.001.

\begin{figure}[t]
	\begin{center}
		\includegraphics[width=\linewidth]{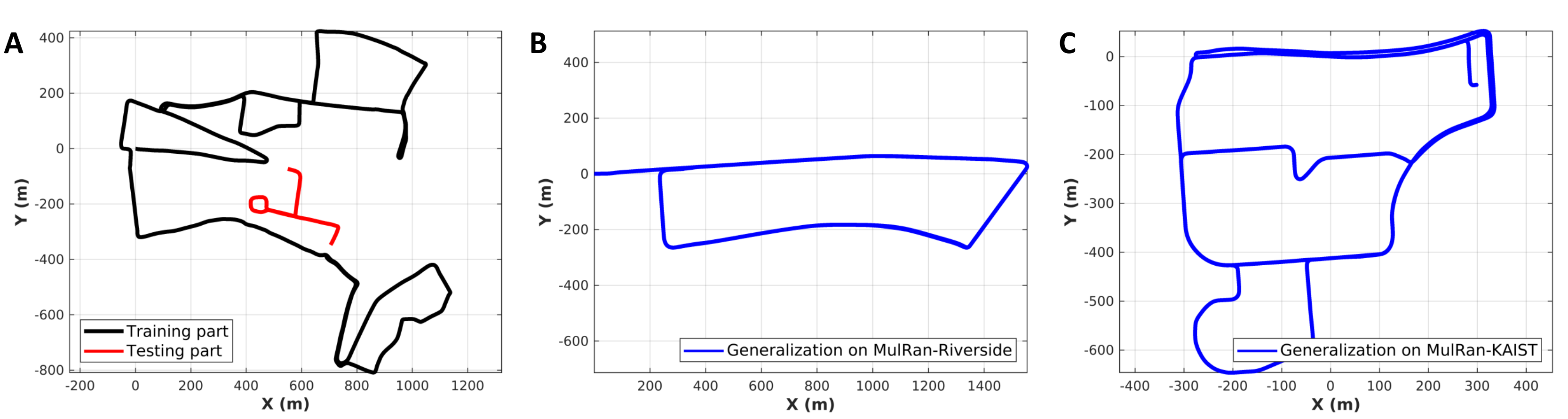}
	\end{center}
	\caption{ \textbf{(A)} We split the trajectory of RobotCar dataset into training and and testing session. \textbf{(B)}\textbf{(C)} The trained model is generalized to the Riverside and KAIST of MulRan dataset for evaluation directly, which contain a driving distance near $7km$ and $6km$ respectively.}
	\label{fig:traj}
\end{figure}

We conduct the experiments on two public dataset, the Oxford Radar RobotCar (RobotCar) dataset \cite{barnes2020oxford, maddern20171} and the Multimodal Range (MulRan) dataset \cite{kim2020mulran}. These two datasets both use the Navtech FMCW radar but the 3D lidar sensors are different, double Velodyne HDL-32E and one Ouster OS1-64 repectively. Our proposed lidar submap construction is able to reduce the differences of the two lidar equipments and settings on mobile robots. To demonstrate the generalization ability, we follow the trainig strategy of our previous work \cite{yin2021rall}, in which only part of RobotCar dataset is used for training. In the test stage, as shown in Figure~\ref{fig:traj}, the learned model is evaluated on another part of the RobotCar, and also generalized to MulRan-Riverside and MulRan-KAIST directly without retraining.

Despite the cross-dataset above, the multi-session evaluation is also used for validation. Specifically, for RobotCar and MulRan datasets, we use one sequence or session as a map database and then select another session on a day as query data. The selected sessions are presented in Table~\ref{sessions}.

To better explore parameter sensitivity, we conduct experiments using various distance thresholds and margin values in $\mathcal{L}_1$. Firstly, we set $m=1.0$ as a constant value, and train the proposed learning model. Then we set different distance thresholds $d$ for evaluation on MulRan-KAIST, which means a found pair of places is considered as true positive when its distance is below $d$ meters. Specifically, we use recall@1 to evaluate the performance, which is calculated as follows: 
\begin{equation} \label{recall@1}
\text{recall@1}=\frac{\text{true positive samples}}{\text{number of query scans}} \%
\end{equation}

\begin{figure}[t]
	\begin{center}
		\includegraphics[width=\linewidth]{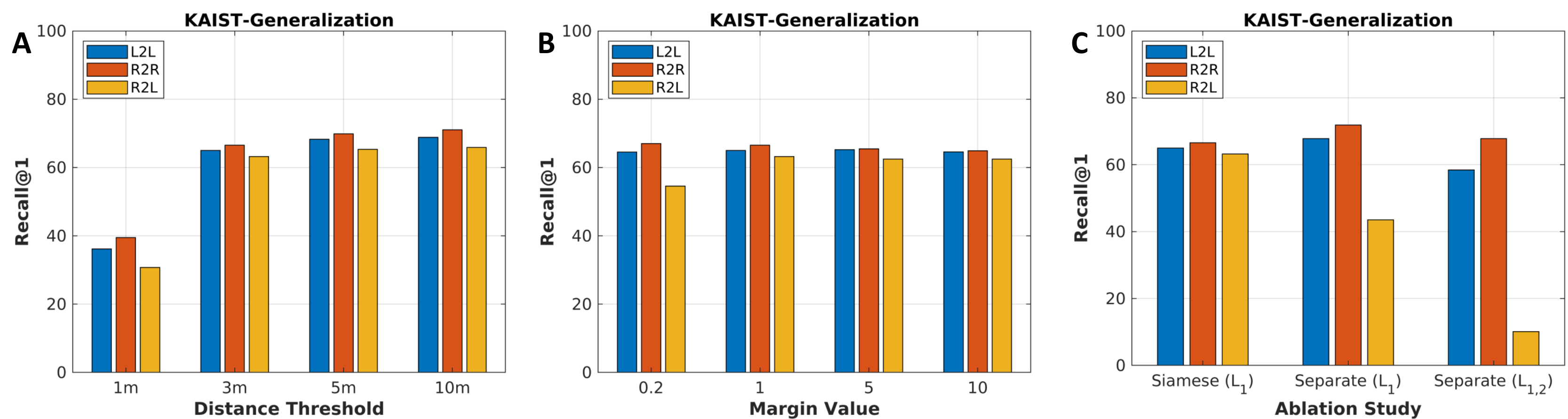}
	\end{center}
	\caption{\textbf{(A)} The parameter sensitivity study of distance thresholds for evaluation. \textbf{(B)} The parameter sensitivity study of margin values for training. \textbf{(C)} The ablation study of proposed framework and loss functions.}
	\label{fig:sensitivity}
\end{figure}

As a result, the sensitivity of distance thresholds is shown in Figure~\ref{fig:sensitivity}(A). The higher threshold is, the better performance the trained model achieves. We select 3 meters as the distance threshold for all the following tests, which was also conducted in our previous work \cite{yin20193d}. Furthermore, we change the margin values and train models respectively. The test result is shown in Figure~\ref{fig:sensitivity}, and our model achieves the best R2L performance when $m=1$.

In this paper, we propose to extract feature embeddings via siamese neural network. One might suggest that the two embeddings of lidar and radar should be achieved with two individual networks. To validate the proposed framework, we conduct ablation study for the framework structure. Firstly, we abandon the siamese network $\mathcal{N}$ and train two separate encoder-decoders via joint learing and loss function $\mathcal{L}_1$. Secondly, we propose to train this new framework with another transformation loss function $\mathcal{L}_2$ together, formulated as follows:
\begin{equation} \label{loss_new}
\mathcal{L}_2 = \frac{1}{\left| \mathcal{F} \right|} \sum_{F\in\mathcal{F}}	\Vert F_R - F_L \Vert
\end{equation}
\begin{equation}
\mathcal{L}_{1,2} = \mathcal{L}_1 + \alpha\mathcal{L}_2
\end{equation}
where $F_R$ and $F_L$ are generated signatures of radar and lidar, and we set $\alpha=0.2$ to balance the triplet loss and transformation loss. Figure~\ref{fig:sensitivity}(C) presents the experimental results with different structures and loss functions. Although the framework with two individual networks performs better than siamese one on R2R and L2L, the shared network matches more correct R2L for heterogeneous place recognition. The transformation loss seems to be redundant for the learning task in this paper. Based on the ablation analysis above, we use the proposed method in Section~\ref{methods} for the following evaluation and comparison.

\begin{table}[!t]
	\begin{center}
		\renewcommand\arraystretch{1.5}
		\caption{Sequences for training and testing}
		\label{sessions}
		\begin{tabular}{p{5cm}<{\centering}|p{1.5cm}<{\centering}p{3cm}<{\centering}p{5cm}<{\centering}}
			\hline
			\hline
			Dataset & Date & Length (km) & Usage \\
			\hline
			Oxford Radar RobotCar & 10/01/2019 & 9.02 & training \& map database \\
			Oxford Radar RobotCar & 11/01/2019 & 9.03 & query data for test \\
			\hline
			MulRan-Riverside & 16/08/2019 & 6.61 & map database \\
			MulRan-Riverside & 02/08/2019 & 7.25 & query data for test \\
			\hline
			MulRan-KAIST & 23/08/2019 & 5.97 & map database \\
			MulRan-KAIST & 20/06/2019 & 6.13 & query data for test \\
			\hline
			\hline
		\end{tabular}
	\end{center}
\end{table}

\subsection{Single-Session: Loop Closure Detection}
Single-session contains one driving or travelling data collected by mobile platform. We evaluate the online place recognition performance on single-session, which equals to the loop closure detection for mapping system. Generally, one single sensor is sufficient to build maps considering the consistency of sensor modality. In this context, radar-to-lidar evaluation is unnecessary, and we perform radar-to-radar on MulRan-KAIST to validate the homogenous place recognition.

\begin{figure}[!t]
	\begin{center}
		\includegraphics[width=\linewidth]{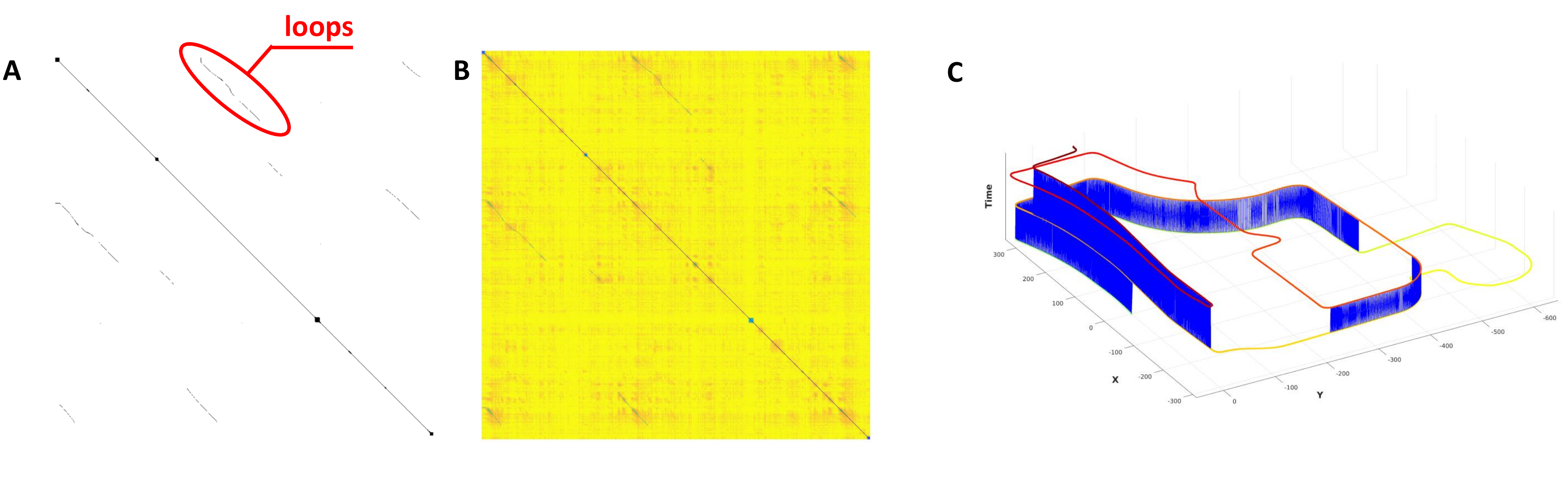}
	\end{center}
	\caption{ \textbf{(A)} The ground truth based binary matrix, where the black points are the loops. \textbf{(B)} The radar-to-radar based similarity matrix generated using the proposed method. \textbf{(C)} The loops closure detection result under certain threshold, and the blue lines are the detected loops. Note that a growing height and a color-changing are added to each pose with respect to the time for visualization.}
	\label{fig:loop}
\end{figure}

We compute similarity matrix \cite{yin2017efficient} and obtain the loop closures under certain threshold. The black points are marked as loop closures in Figure~\ref{fig:loop}(A) and the darker pixels are with higher probabilities to be loop closures in Figure~\ref{fig:loop}(B). It is clear that there are true positive loops in the similarity matrix, and a number of loops can be found via thresholding. In Figure~\ref{fig:loop}(C), the visualization result shows that our proposed method is able to detect radar loop closures when the vehicle is driving in opposite directions. Overall, the qualitative result of online place recognition demonstrates that our proposed method is feasible for consistent map building. While for the global localization, multi-session place recognition is required, and we conduct quantitive experiments as follows.

\subsection{Multi-Session: Global Localization}


In this sub-section, we evaluate the place recognition performance on multi-session data. Specifically, the first session is used as the map database, another session is regarded as the query input, thus achieveing the global localization across days.

The proposed joint learning based method is compared to other two competitive methods. First, the ScanContext is used for comparison, which achieves not only 3D lidar place recognition \cite{kim2018scan} but also 2D radar place recognition in the recent publication \cite{kim2020mulran}. Secondly, the DiSCO method \cite{xu2020disco} is implemented as another comparison, and the quadruplet loss term is used in the learning stage. DiSCO is not designed for heterogeneuos place recognition, so we train two models for L2L and R2R separately, and test the R2L using the signatures from these models. On the other hand, DiSCO can be regarded as the model without the joint learning in this paper. Finally, for making a fair comparison, we use the lidar submaps as input for all the methods in this paper. 

\begin{figure}[t]
	\begin{center}
		\includegraphics[width=\linewidth]{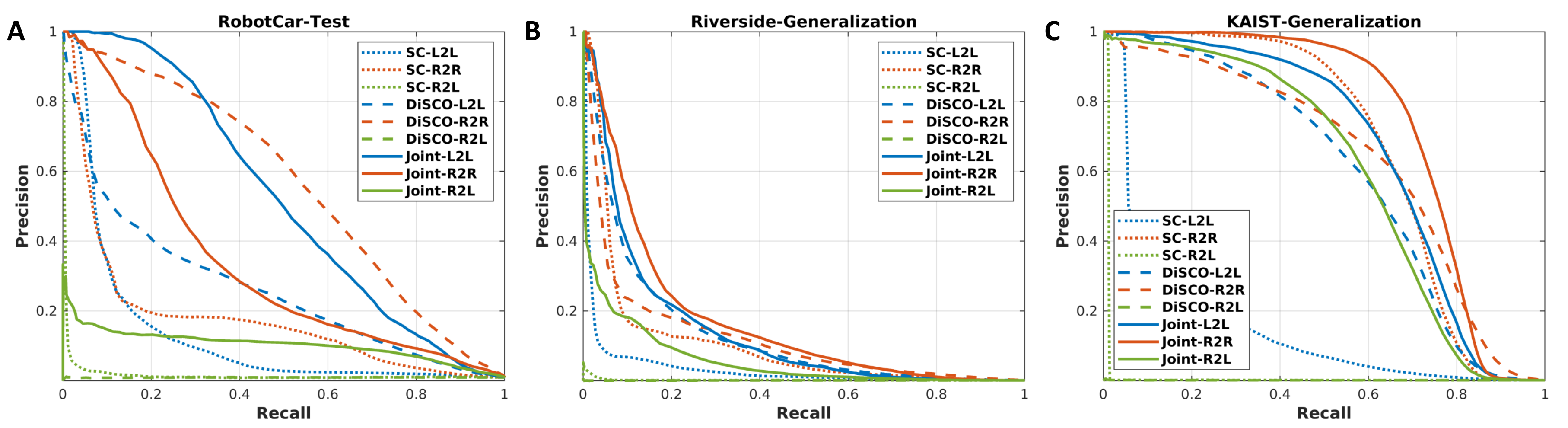}
	\end{center}
	\caption{ \textbf{(A)} The precision-recall curves of the RobotCar testing session. \textbf{(B)}\textbf{(C)} The model trained from RobotCar dataset is generalized to MulRan-Riverside and MulRan-KAIST respectively. }
	\label{fig:pr}
\end{figure}

Figure \ref{fig:pr} presents the precision-recall curves, which are generated from the similarity matrices compared to the ground truth based binary matrices. Since the computing of similarity matrix is much more time consuming for ScanContext, we only present the precision-recall curves for DiSCO and our proposed method. In addition, we also provide maximum $F_1$ scores in Table \ref{Fscore}, and recall@1 results in Table \ref{recall}, which is based on how many correct top-1 can be found using place recognition methods. In \cite{kim2018scan, kim2020mulran}, the top-1 is searched with a coarse-to-fine strategy, and we set the number of coarse candidates as 1\% of the database.

As a result, in Figure \ref{fig:pr} and Table \ref{recall}, our proposed method achieves comparable results on R2R and L2L, and also on R2L application, which is the evaluation result based on lidar database and radar query. As for ScanContext and DiSCO, both two methods achieve high performance on L2L and R2R, but radar and lidar are not connected to each other in these methods, resulting in much lower performance on R2L. We also note that ScanContext performs much worse with MulRan-KAIST, in which many dynamical objects exist. The other two learning-based methods are able to handle these challenging environments. Overall, the multi-session place recognition results demonstrate that our proposed method achieves both homogeneous and heterogeneuos place recognition, and our model requires less training stage compared to DiSCO. 

\begin{table}[!t]
	\begin{center}
		\renewcommand\arraystretch{1.5}
		\caption{maximum $F_1$ score of precision-recall curves}
		\label{Fscore}
		\begin{tabular}{p{5.5cm}<{\centering}|p{0.9cm}<{\centering}p{0.9cm}<{\centering}p{0.9cm}<{\centering}p{0.9cm}<{\centering}p{0.9cm}<{\centering}p{0.9cm}<{\centering}p{0.9cm}<{\centering}p{0.9cm}<{\centering}p{0.9cm}<{\centering}}
			\hline
			\hline
			\multirow{2}*{Dataset} &\multicolumn{3}{c}{RobotCar-Test} &\multicolumn{3}{c}{MulRan-Riverside} &\multicolumn{3}{c}{MulRan-KAIST}  \\
			~ & L2L & R2R & R2L & L2L & R2R & R2L & L2L & R2R & R2L \\
			\hline
			ScanContext \cite{kim2020mulran} & 0.18 & 0.24 & 0.04 & 0.08 & 0.16 & 0.02 & 0.24 & 0.67 & 0.02 \\
			DiSCO \cite{xu2020disco}  & 0.33 & \textbf{0.56} & 0.02 & 0.20& 0.20& 0.00 & 0.59 & 0.63 & 0.01 \\
			Joint Learning  & \textbf{0.50} & 0.35 & \textbf{0.18} & \textbf{0.21} & \textbf{0.22} &\textbf{0.14} & \textbf{0.66} & \textbf{0.74} & \textbf{0.61} \\
			\hline
			\hline
		\end{tabular}
	\end{center}
\end{table}

\begin{table}[!t]
	\begin{center}
		\renewcommand\arraystretch{1.5}
		\caption{Recall@1 (\%) of multi-session place recognition}
		\label{recall}
		\begin{tabular}{p{5.5cm}<{\centering}|p{0.9cm}<{\centering}p{0.9cm}<{\centering}p{0.9cm}<{\centering}p{0.9cm}<{\centering}p{0.9cm}<{\centering}p{0.9cm}<{\centering}p{0.9cm}<{\centering}p{0.9cm}<{\centering}p{0.9cm}<{\centering}}
			\hline
			\hline
			\multirow{2}*{Dataset} &\multicolumn{3}{c}{RobotCar-Test} &\multicolumn{3}{c}{MulRan-Riverside} &\multicolumn{3}{c}{MulRan-KAIST}  \\
			~ & L2L & R2R & R2L & L2L & R2R & R2L & L2L & R2R & R2L \\
			\hline
			ScanContext \cite{kim2020mulran} & 92.51 &91.35 & 1.33 & 37.22& 42.68& 1.17& 28.81 & \textbf{72.65} & 0.55 \\
			DiSCO \cite{xu2020disco}  & 89.68 & 93.01 & 1.66 & \textbf{42.19}& \textbf{44.08}& 0.99 & \textbf{66.77} & 70.86 & 0.33 \\
			Joint Learning  & \textbf{93.68} & \textbf{93.18} & \textbf{61.23} & 38.89& 41.15 &\textbf{26.20} & 65.01 & 66.56 & \textbf{63.22} \\
			\hline
			\hline
		\end{tabular}
	\end{center}
\end{table}

\subsection{Case Study}
\label{case_study}

\begin{figure}[t]
	\begin{center}
		\includegraphics[width=\linewidth]{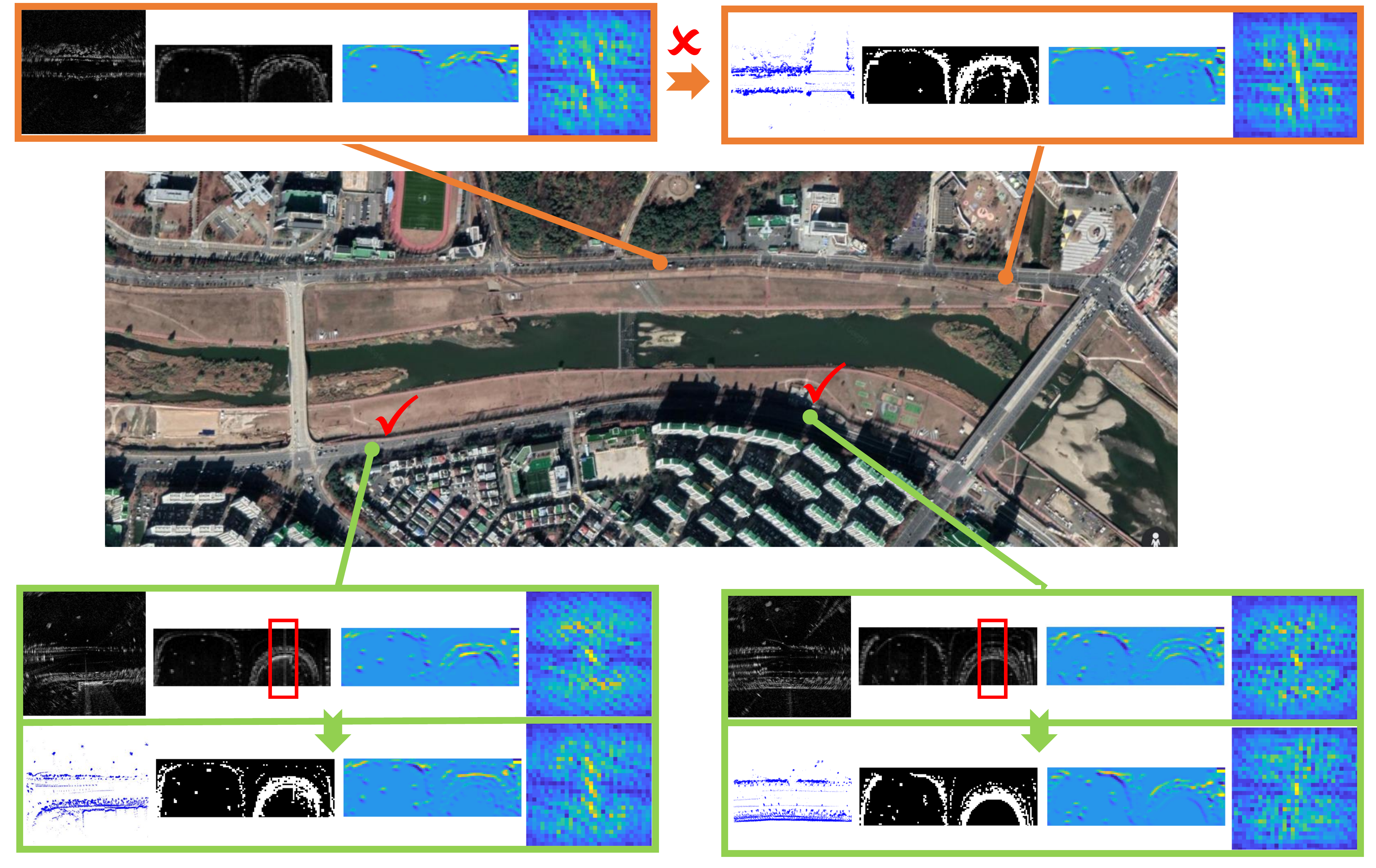}
	\end{center}
	\caption{Case study examples on radar-to-lidar place recognition, where the lidar database and radar query are collected in different days. We also present ScanContext representations, feature embeddings and final signatures respectively. Some false positives by saturation are also marked in red boxes.}
	\label{fig:case}
\end{figure}

Finally, we present several case study examples on the challenging MulRan-Riverside, where many structural features are repetitive along the road. As shown in Figure \ref{fig:case}, the trained model results in a failed case since the bushes and buildings are quite similar at two different places, and in this context, the radar from the query data is matched to the wrong lidar based database. As for the two correct cases, there exist specific features on the sides of streets, corners and buildings etc., thus making the place recognition model more robust in this challenging environments.

In Figure~\ref{fig:case}, it is obvious that the semantics near the roads are kept in the feature embeddings, which are enhanced via the joint learning in this paper. We also note that there are false positive noises by saturation (in red boxes), but the noises are removed in the learned feature embeddings, thus demonstrating the effectiveness of the proposed joint learning paradigm.

\section{Conclusion}
\label{conclusion}

In this paper, we propose to train a joint learning system for radar and lidar place recognition, which helps the robot recognize the revisted lidar submaps using current radar scan. Specifically, we first encode the radar and lidar points based on ScanContext, then we build a shared U-Net to transform the handcrafted features to the learned representations. To achieve the place recognition, we also apply the triplet loss as the supervision. The whole system is trained jointly with both lidar and radar input. Finally, we conduct the training and testing on the public Oxford RobotCar dataset, and also the generalization on MulRan dataset. Compared to the existing place recognition methods, our proposed framework achieves not only single-sensor based place recognition, but also the heterogeneous place recognition of radar-to-lidar, demonstrating the effectiveness of our proposed joint learning framework.

Despite the conclusions above, we also consider there still remain several promising directions for heterogeneous measurements based robotic perception. Firstly, the submap building is critical for the heterogeneous place recognition in this paper, which can be improved with a more informative method \cite{adolfsson2019submap}. Secondly, we consider a GPS-aided or sequential-based place recognition is desired for real applications, thus making the perception system more efficient and effective in the time domain. Finally, we consider the integration of place recognition and pose estimator, Monte Carlo localization for example \cite{yin20193d, sun2020localising}, is a good choice for metric robot localization.

\section*{Conflict of Interest Statement}

The authors declare that the research was conducted in the absence of any commercial or financial relationships that could be construed as a potential conflict of interest.

\section*{Author Contributions}
HY: methodology, implementation, experiment design, visualization, and writing. XX: methodology and review. YW: analysis, supervision, and review. RX: supervision and project administration. All authors contributed to the article and approved the submitted version.

\section*{Funding}

This work was supported by the National Key R\&D Program of China under grant 2020YFB1313300.

\section*{Acknowledgments}
We sincerely thank the reviewers for the careful reading and the comments. We also thank the public RobotCar and MulRan datasets.
%

\section*{Data Availability Statement}
The original contributions presented in the study are included in the article/supplementary material, further inquiries can be directed to the corresponding author/s.

\bibliographystyle{frontiersinSCNS_ENG_HUMS} 
\bibliography{test}

\begin{thebibliography}{48}
\providecommand{\natexlab}[1]{#1}
\expandafter\ifx\csname urlstyle\endcsname\relax
  \providecommand{\doi}[1]{doi:\discretionary{}{}{}#1}\else
  \providecommand{\doi}{doi:\discretionary{}{}{}\begingroup
  \urlstyle{rm}\Url}\fi
\providecommand{\selectlanguage}[1]{\relax}
\providecommand{\bibAnnoteFile}[1]{%
  \IfFileExists{#1}{\begin{quotation}\noindent\textsc{Key:} #1\\
  \textsc{Annotation:}\ \input{#1}\end{quotation}}{}}
\providecommand{\bibAnnote}[2]{%
  \begin{quotation}\noindent\textsc{Key:} #1\\
  \textsc{Annotation:}\ #2\end{quotation}}

\bibitem[{Adolfsson et~al.(2019)Adolfsson, Lowry, Magnusson, Lilienthal, and
  Andreasson}]{adolfsson2019submap}
Adolfsson, D., Lowry, S., Magnusson, M., Lilienthal, A., and Andreasson, H.
  (2019).
\newblock A submap per perspective-selecting subsets for super mapping that
  afford superior localization quality.
\newblock In \emph{2019 European Conference on Mobile Robots (ECMR)} (IEEE),
  1--7
\bibAnnoteFile{adolfsson2019submap}

\bibitem[{Arandjelovic et~al.(2016)Arandjelovic, Gronat, Torii, Pajdla, and
  Sivic}]{arandjelovic2016netvlad}
Arandjelovic, R., Gronat, P., Torii, A., Pajdla, T., and Sivic, J. (2016).
\newblock Netvlad: Cnn architecture for weakly supervised place recognition.
\newblock In \emph{Proceedings of the IEEE conference on computer vision and
  pattern recognition}. 5297--5307
\bibAnnoteFile{arandjelovic2016netvlad}

\bibitem[{Barnes et~al.(2020{\natexlab{a}})Barnes, Gadd, Murcutt, Newman, and
  Posner}]{barnes2020oxford}
Barnes, D., Gadd, M., Murcutt, P., Newman, P., and Posner, I.
  (2020{\natexlab{a}}).
\newblock The oxford radar robotcar dataset: A radar extension to the oxford
  robotcar dataset.
\newblock In \emph{2020 IEEE International Conference on Robotics and
  Automation (ICRA)} (IEEE), 6433--6438
\bibAnnoteFile{barnes2020oxford}

\bibitem[{Barnes et~al.(2020{\natexlab{b}})Barnes, Weston, and
  Posner}]{barnes2020masking}
Barnes, D., Weston, R., and Posner, I. (2020{\natexlab{b}}).
\newblock Masking by moving: Learning distraction-free radar odometry from pose
  information.
\newblock In \emph{Conference on Robot Learning} (PMLR), 303--316
\bibAnnoteFile{barnes2020masking}

\bibitem[{Bay et~al.(2006)Bay, Tuytelaars, and Van~Gool}]{bay2006surf}
Bay, H., Tuytelaars, T., and Van~Gool, L. (2006).
\newblock Surf: Speeded up robust features.
\newblock In \emph{European conference on computer vision} (Springer), 404--417
\bibAnnoteFile{bay2006surf}

\bibitem[{Bosse and Zlot(2013)}]{bosse2013place}
Bosse, M. and Zlot, R. (2013).
\newblock Place recognition using keypoint voting in large 3d lidar datasets.
\newblock In \emph{2013 IEEE International Conference on Robotics and
  Automation} (IEEE), 2677--2684
\bibAnnoteFile{bosse2013place}

\bibitem[{Carballo et~al.(2020)Carballo, Lambert, Monrroy, Wong, Narksri,
  Kitsukawa et~al.}]{carballo2020libre}
Carballo, A., Lambert, J., Monrroy, A., Wong, D., Narksri, P., Kitsukawa, Y.,
  et~al. (2020).
\newblock Libre: The multiple 3d lidar dataset.
\newblock In \emph{2020 IEEE Intelligent Vehicles Symposium (IV)} (IEEE)
\bibAnnoteFile{carballo2020libre}

\bibitem[{Cattaneo et~al.(2020)Cattaneo, Vaghi, Fontana, Ballardini, and
  Sorrenti}]{cattaneo2020global}
Cattaneo, D., Vaghi, M., Fontana, S., Ballardini, A.~L., and Sorrenti, D.~G.
  (2020).
\newblock Global visual localization in lidar-maps through shared 2d-3d
  embedding space.
\newblock In \emph{2020 IEEE International Conference on Robotics and
  Automation (ICRA)} (IEEE), 4365--4371
\bibAnnoteFile{cattaneo2020global}

\bibitem[{Cen and Newman(2018)}]{cen2018precise}
Cen, S.~H. and Newman, P. (2018).
\newblock Precise ego-motion estimation with millimeter-wave radar under
  diverse and challenging conditions.
\newblock In \emph{2018 IEEE International Conference on Robotics and
  Automation (ICRA)} (IEEE), 6045--6052
\bibAnnoteFile{cen2018precise}

\bibitem[{Chen et~al.(2020)Chen, L{\"a}be, Milioto, R{\"o}hling, Vysotska, Haag
  et~al.}]{chen2020overlapnet}
Chen, X., L{\"a}be, T., Milioto, A., R{\"o}hling, T., Vysotska, O., Haag, A.,
  et~al. (2020).
\newblock Overlapnet: Loop closing for lidar-based slam.
\newblock In \emph{Proc. of Robotics: Science and Systems (RSS)}
\bibAnnoteFile{chen2020overlapnet}

\bibitem[{Cummins and Newman(2008)}]{cummins2008fab}
Cummins, M. and Newman, P. (2008).
\newblock Fab-map: Probabilistic localization and mapping in the space of
  appearance.
\newblock \emph{The International Journal of Robotics Research} 27, 647--665
\bibAnnoteFile{cummins2008fab}

\bibitem[{Ding et~al.(2019)Ding, Wang, Xiong, Li, Tang, Yin
  et~al.}]{ding2019persistent}
Ding, X., Wang, Y., Xiong, R., Li, D., Tang, L., Yin, H., et~al. (2019).
\newblock Persistent stereo visual localization on cross-modal invariant map.
\newblock \emph{IEEE Transactions on Intelligent Transportation Systems}
\bibAnnoteFile{ding2019persistent}

\bibitem[{Dub{\'e} et~al.(2020)Dub{\'e}, Cramariuc, Dugas, Sommer, Dymczyk,
  Nieto et~al.}]{dube2020segmap}
Dub{\'e}, R., Cramariuc, A., Dugas, D., Sommer, H., Dymczyk, M., Nieto, J.,
  et~al. (2020).
\newblock Segmap: Segment-based mapping and localization using data-driven
  descriptors.
\newblock \emph{The International Journal of Robotics Research} 39, 339--355
\bibAnnoteFile{dube2020segmap}

\bibitem[{Elhousni and Huang(2020)}]{elhousni2020survey}
Elhousni, M. and Huang, X. (2020).
\newblock A survey on 3d lidar localization for autonomous vehicles.
\newblock In \emph{2020 IEEE Intelligent Vehicles Symposium (IV)} (IEEE),
  1879--1884
\bibAnnoteFile{elhousni2020survey}

\bibitem[{Feng et~al.(2019)Feng, Hu, Ang, and Lee}]{feng20192d3d}
Feng, M., Hu, S., Ang, M.~H., and Lee, G.~H. (2019).
\newblock 2d3d-matchnet: Learning to match keypoints across 2d image and 3d
  point cloud.
\newblock In \emph{2019 International Conference on Robotics and Automation
  (ICRA)} (IEEE), 4790--4796
\bibAnnoteFile{feng20192d3d}

\bibitem[{Filliat(2007)}]{filliat2007visual}
Filliat, D. (2007).
\newblock A visual bag of words method for interactive qualitative localization
  and mapping.
\newblock In \emph{Proceedings 2007 IEEE International Conference on Robotics
  and Automation} (IEEE), 3921--3926
\bibAnnoteFile{filliat2007visual}

\bibitem[{Gadd et~al.(2020)Gadd, De~Martini, and Newman}]{gadd2020look}
Gadd, M., De~Martini, D., and Newman, P. (2020).
\newblock Look around you: Sequence-based radar place recognition with learned
  rotational invariance.
\newblock In \emph{2020 IEEE/ION Position, Location and Navigation Symposium
  (PLANS)}. 270--276
\bibAnnoteFile{gadd2020look}

\bibitem[{Gentil et~al.(2020)Gentil, Vayugundla, Giubilato, St{\"u}rzl,
  Vidal-Calleja, and Triebel}]{gentil2020gaussian}
Gentil, C.~L., Vayugundla, M., Giubilato, R., St{\"u}rzl, W., Vidal-Calleja,
  T., and Triebel, R. (2020).
\newblock Gaussian process gradient maps for loop-closure detection in
  unstructured planetary environments.
\newblock \emph{arXiv preprint arXiv:2009.00221}
\bibAnnoteFile{gentil2020gaussian}

\bibitem[{Granstr{\"o}m et~al.(2011)Granstr{\"o}m, Sch{\"o}n, Nieto, and
  Ramos}]{granstrom2011learning}
Granstr{\"o}m, K., Sch{\"o}n, T.~B., Nieto, J.~I., and Ramos, F.~T. (2011).
\newblock Learning to close loops from range data.
\newblock \emph{The international journal of robotics research} 30, 1728--1754
\bibAnnoteFile{granstrom2011learning}

\bibitem[{He et~al.(2016)He, Wang, and Zhang}]{he2016m2dp}
He, L., Wang, X., and Zhang, H. (2016).
\newblock M2dp: A novel 3d point cloud descriptor and its application in loop
  closure detection.
\newblock In \emph{2016 IEEE/RSJ International Conference on Intelligent Robots
  and Systems (IROS)} (IEEE), 231--237
\bibAnnoteFile{he2016m2dp}

\bibitem[{Hong et~al.(2020)Hong, Petillot, and Wang}]{hong2020radarslam}
Hong, Z., Petillot, Y., and Wang, S. (2020).
\newblock Radarslam: Radar based large-scale slam in all weathers.
\newblock In \emph{2020 IEEE/RSJ International Conference on Intelligent Robots
  and Systems (IROS)}
\bibAnnoteFile{hong2020radarslam}

\bibitem[{J{\'e}gou et~al.(2010)J{\'e}gou, Douze, Schmid, and
  P{\'e}rez}]{jegou2010aggregating}
J{\'e}gou, H., Douze, M., Schmid, C., and P{\'e}rez, P. (2010).
\newblock Aggregating local descriptors into a compact image representation.
\newblock In \emph{2010 IEEE computer society conference on computer vision and
  pattern recognition} (IEEE), 3304--3311
\bibAnnoteFile{jegou2010aggregating}

\bibitem[{Kim and Kim(2018)}]{kim2018scan}
Kim, G. and Kim, A. (2018).
\newblock Scan context: Egocentric spatial descriptor for place recognition
  within 3d point cloud map.
\newblock In \emph{2018 IEEE/RSJ International Conference on Intelligent Robots
  and Systems (IROS)} (IEEE), 4802--4809
\bibAnnoteFile{kim2018scan}

\bibitem[{Kim et~al.(2020)Kim, Park, Cho, Jeong, and Kim}]{kim2020mulran}
Kim, G., Park, Y.~S., Cho, Y., Jeong, J., and Kim, A. (2020).
\newblock Mulran: Multimodal range dataset for urban place recognition.
\newblock In \emph{IEEE International Conference on Robotics and Automation
  (ICRA)}
\bibAnnoteFile{kim2020mulran}

\bibitem[{Kingma and Ba(2014)}]{kingma2014adam}
Kingma, D.~P. and Ba, J. (2014).
\newblock Adam: A method for stochastic optimization.
\newblock \emph{arXiv preprint arXiv:1412.6980}
\bibAnnoteFile{kingma2014adam}

\bibitem[{Krstanovic et~al.(2012)Krstanovic, Keller, and
  Groft}]{krstanovic2012radar}
[Dataset] Krstanovic, C., Keller, S., and Groft, E. (2012).
\newblock Radar vehicle detection system.
\newblock US Patent 8,279,107
\bibAnnoteFile{krstanovic2012radar}

\bibitem[{Latif et~al.(2018)Latif, Garg, Milford, and
  Reid}]{latif2018addressing}
Latif, Y., Garg, R., Milford, M., and Reid, I. (2018).
\newblock Addressing challenging place recognition tasks using generative
  adversarial networks.
\newblock In \emph{2018 IEEE International Conference on Robotics and
  Automation (ICRA)} (IEEE), 2349--2355
\bibAnnoteFile{latif2018addressing}

\bibitem[{Li et~al.(2017)Li, Yang, Wang, and Wang}]{li2017overview}
Li, L., Yang, M., Wang, B., and Wang, C. (2017).
\newblock An overview on sensor map based localization for automated driving.
\newblock In \emph{2017 Joint Urban Remote Sensing Event (JURSE)} (IEEE), 1--4
\bibAnnoteFile{li2017overview}

\bibitem[{Liu et~al.(2019)Liu, Zhou, Suo, Yin, Chen, Wang et~al.}]{liu2019lpd}
Liu, Z., Zhou, S., Suo, C., Yin, P., Chen, W., Wang, H., et~al. (2019).
\newblock Lpd-net: 3d point cloud learning for large-scale place recognition
  and environment analysis.
\newblock In \emph{Proceedings of the IEEE/CVF International Conference on
  Computer Vision}. 2831--2840
\bibAnnoteFile{liu2019lpd}

\bibitem[{Lowry et~al.(2015)Lowry, S{\"u}nderhauf, Newman, Leonard, Cox, Corke
  et~al.}]{lowry2015visual}
Lowry, S., S{\"u}nderhauf, N., Newman, P., Leonard, J.~J., Cox, D., Corke, P.,
  et~al. (2015).
\newblock Visual place recognition: A survey.
\newblock \emph{IEEE Transactions on Robotics} 32, 1--19
\bibAnnoteFile{lowry2015visual}

\bibitem[{Maddern et~al.(2017)Maddern, Pascoe, Linegar, and
  Newman}]{maddern20171}
Maddern, W., Pascoe, G., Linegar, C., and Newman, P. (2017).
\newblock 1 year, 1000 km: The oxford robotcar dataset.
\newblock \emph{The International Journal of Robotics Research} 36, 3--15
\bibAnnoteFile{maddern20171}

\bibitem[{Milford and Wyeth(2012)}]{milford2012seqslam}
Milford, M.~J. and Wyeth, G.~F. (2012).
\newblock Seqslam: Visual route-based navigation for sunny summer days and
  stormy winter nights.
\newblock In \emph{2012 IEEE international conference on robotics and
  automation} (IEEE), 1643--1649
\bibAnnoteFile{milford2012seqslam}

\bibitem[{Pan et~al.(2020)Pan, Xu, Li, Wang, and Xiong}]{pan2020coral}
Pan, Y., Xu, X., Li, W., Wang, Y., and Xiong, R. (2020).
\newblock Coral: Colored structural representation for bi-modal place
  recognition.
\newblock \emph{arXiv preprint arXiv:2011.10934}
\bibAnnoteFile{pan2020coral}

\bibitem[{Paszke et~al.(2019)Paszke, Gross, Massa, Lerer, Bradbury, Chanan
  et~al.}]{paszke2019pytorch}
Paszke, A., Gross, S., Massa, F., Lerer, A., Bradbury, J., Chanan, G., et~al.
  (2019).
\newblock Pytorch: An imperative style, high-performance deep learning library.
\newblock In \emph{Advances in neural information processing systems}.
  8026--8037
\bibAnnoteFile{paszke2019pytorch}

\bibitem[{Ronneberger et~al.(2015)Ronneberger, Fischer, and
  Brox}]{ronneberger2015u}
Ronneberger, O., Fischer, P., and Brox, T. (2015).
\newblock U-net: Convolutional networks for biomedical image segmentation.
\newblock In \emph{International Conference on Medical image computing and
  computer-assisted intervention} (Springer), 234--241
\bibAnnoteFile{ronneberger2015u}

\bibitem[{Rublee et~al.(2011)Rublee, Rabaud, Konolige, and
  Bradski}]{rublee2011orb}
Rublee, E., Rabaud, V., Konolige, K., and Bradski, G. (2011).
\newblock Orb: An efficient alternative to sift or surf.
\newblock In \emph{2011 International conference on computer vision} (Ieee),
  2564--2571
\bibAnnoteFile{rublee2011orb}

\bibitem[{S{\u{a}}ftescu et~al.(2020)S{\u{a}}ftescu, Gadd, De~Martini, Barnes,
  and Newman}]{suaftescu2020kidnapped}
S{\u{a}}ftescu, {\c{S}}., Gadd, M., De~Martini, D., Barnes, D., and Newman, P.
  (2020).
\newblock Kidnapped radar: Topological radar localisation using
  rotationally-invariant metric learning.
\newblock In \emph{2020 IEEE International Conference on Robotics and
  Automation (ICRA)} (IEEE), 4358--4364
\bibAnnoteFile{suaftescu2020kidnapped}

\bibitem[{Sun et~al.(2020)Sun, Adolfsson, Magnusson, Andreasson, Posner, and
  Duckett}]{sun2020localising}
Sun, L., Adolfsson, D., Magnusson, M., Andreasson, H., Posner, I., and Duckett,
  T. (2020).
\newblock Localising faster: Efficient and precise lidar-based robot
  localisation in large-scale environments.
\newblock In \emph{2020 IEEE International Conference on Robotics and
  Automation (ICRA)} (IEEE), 4386--4392
\bibAnnoteFile{sun2020localising}

\bibitem[{Tang et~al.(2020)Tang, De~Martini, Barnes, and Newman}]{tang2020rsl}
Tang, T.~Y., De~Martini, D., Barnes, D., and Newman, P. (2020).
\newblock Rsl-net: Localising in satellite images from a radar on the ground.
\newblock \emph{IEEE Robotics and Automation Letters} 5, 1087--1094
\bibAnnoteFile{tang2020rsl}

\bibitem[{Uy and Lee(2018)}]{uy2018pointnetvlad}
Uy, M.~A. and Lee, G.~H. (2018).
\newblock Pointnetvlad: Deep point cloud based retrieval for large-scale place
  recognition.
\newblock In \emph{Proceedings of the IEEE Conference on Computer Vision and
  Pattern Recognition}. 4470--4479
\bibAnnoteFile{uy2018pointnetvlad}

\bibitem[{Wang et~al.(2019)Wang, Sun, Xu, Sarma, Yang, and
  Kong}]{wang2019lidar}
Wang, Y., Sun, Z., Xu, C.-Z., Sarma, S., Yang, J., and Kong, H. (2019).
\newblock Lidar iris for loop-closure detection.
\newblock \emph{arXiv preprint arXiv:1912.03825}
\bibAnnoteFile{wang2019lidar}

\bibitem[{Xie et~al.(2020)Xie, Pan, Peng, Liu, and Ying}]{xie2020large}
Xie, S., Pan, C., Peng, Y., Liu, K., and Ying, S. (2020).
\newblock Large-scale place recognition based on camera-lidar fused descriptor.
\newblock \emph{Sensors} 20, 2870
\bibAnnoteFile{xie2020large}

\bibitem[{Xu et~al.(2020)Xu, Yin, Chen, Wang, and Xiong}]{xu2020disco}
Xu, X., Yin, H., Chen, Z., Wang, Y., and Xiong, R. (2020).
\newblock Disco: Differentiable scan context with orientation.
\newblock \emph{arXiv preprint arXiv:2010.10949}
\bibAnnoteFile{xu2020disco}

\bibitem[{Yin et~al.(2021)Yin, Chen, Wang, and Xiong}]{yin2021rall}
Yin, H., Chen, R., Wang, Y., and Xiong, R. (2021).
\newblock Rall: End-to-end radar localization on lidar map using differentiable
  measurement model.
\newblock \emph{IEEE Transactions on Intelligent Transportation Systems}
\bibAnnoteFile{yin2021rall}

\bibitem[{Yin et~al.(2017)Yin, Ding, Tang, Wang, and Xiong}]{yin2017efficient}
Yin, H., Ding, X., Tang, L., Wang, Y., and Xiong, R. (2017).
\newblock Efficient 3d lidar based loop closing using deep neural network.
\newblock In \emph{2017 IEEE International Conference on Robotics and
  Biomimetics (ROBIO)} (IEEE), 481--486
\bibAnnoteFile{yin2017efficient}

\bibitem[{Yin et~al.(2019)Yin, Wang, Ding, Tang, Huang, and Xiong}]{yin20193d}
Yin, H., Wang, Y., Ding, X., Tang, L., Huang, S., and Xiong, R. (2019).
\newblock 3d lidar-based global localization using siamese neural network.
\newblock \emph{IEEE Transactions on Intelligent Transportation Systems} 21,
  1380--1392
\bibAnnoteFile{yin20193d}

\bibitem[{Yin et~al.(2020)Yin, Wang, Tang, and Xiong}]{yin2020radar}
Yin, H., Wang, Y., Tang, L., and Xiong, R. (2020).
\newblock Radar-on-lidar: metric radar localization on prior lidar maps.
\newblock In \emph{2020 IEEE International Conference on Real-time Computing
  and Robotics (RCAR)} (IEEE).
\newblock Best Conference Paper Award
\bibAnnoteFile{yin2020radar}

\bibitem[{Zhang et~al.(2020)Zhang, Wang, and Su}]{zhang2020visual}
Zhang, X., Wang, L., and Su, Y. (2020).
\newblock Visual place recognition: A survey from deep learning perspective.
\newblock \emph{Pattern Recognition} , 107760
\bibAnnoteFile{zhang2020visual}

\end{thebibliography}

\end{document}